\documentclass[letterpaper, 10 pt, conference]{ieeeconf}  

\IEEEoverridecommandlockouts                              

\usepackage{graphics} 
\usepackage{epsfig} 
\usepackage{amsmath} 
\usepackage{amssymb}  
\usepackage{graphicx}      
\usepackage{booktabs}      
\usepackage[table]{xcolor} 
\usepackage{arydshln}      
\usepackage{multirow}      
\usepackage{pifont}
\usepackage{newtxtext, newtxmath}

\title{\LARGE \bf
VCS-SLAM: Visibility-Consistent Semantic Fusion for 3D Gaussian SLAM
}

\author{Raman Jha$^{1,2,4}$, Shuaihang Yuan$^{2,3\dagger}$, and Yi Fang$^{1,2,3}$%
\thanks{$^{1}$New York University, Electrical \& Computer Engineering Dept., Brooklyn, NY 11201, USA.}%
\thanks{$^{2}$Embodied AI and Robotics (AIR) Lab, NYU Abu Dhabi, UAE.}%
\thanks{$^{3}$NYUAD Center for Artificial Intelligence and Robotics (CAIR), Abu Dhabi, UAE.}%
\thanks{$^{4}$New York University Abu Dhabi, Electrical Engineering, Abu Dhabi 129188, UAE.}%
\thanks{$^{\dagger}$Corresponding author.}%
}

\begin{document}

\maketitle
\thispagestyle{empty}
\pagestyle{empty}

\begin{abstract}
Semantic 3D Gaussian SLAM commonly fuses 2D semantic priors into a persistent 3D map using uniform optimization weights. However, such priors are not equally reliable in online mapping: occlusions, unsupported semantic boundaries, and ambiguous ray geometry introduce persistent semantic artifacts into the global Gaussian map. We propose VCS-SLAM, a visibility-consistent semantic fusion framework for RGB-D 3D Gaussian SLAM. Instead of treating all semantic priors as uniformly valid signals, VCS-SLAM evaluates their geometric reliability through visibility consistency, surface-supported boundary evidence, and ray-level conflict uncertainty. The resulting reliability-aware objective suppresses occluded semantic updates, reduces unsupported semantic bleeding, and delays premature label assignment in ambiguous regions. Experiments on Replica show improved semantic consistency, boundary preservation, and reconstruction quality, while ScanNet results confirm competitive tracking under real RGB-D input. Under structured corruption of the 2D priors, VCS-SLAM degrades at little more than half the rate of a comparable semantic Gaussian SLAM baseline and outperforms naive gating, smoothness, and entropy-weighting variants of its own modules, indicating that the gains follow from the specific formulation rather than from reweighting alone.
Code is available at \textit{https://github.com/jhakrraman/VCS-SLAM}
\end{abstract}


\section{INTRODUCTION}

Visual SLAM fundamentally challenges robotics by requiring simultaneous camera pose estimation and incremental environment mapping \cite{6162880}. While traditional frameworks perform robustly in static settings, their performance deteriorates during complex real-world applications. Although multi-modal sensor fusion addresses visual degradation in extreme conditions \cite{11084466}, ensuring spatial and geometric consistency of semantic representations in 3D environments remains a critical bottleneck. Integrating visibility consistency and surface coupling improves structural mapping and object association, providing richer representations for advanced deployments.

\begin{figure}[htbp]
    \vspace{-14pt}
    \centering
    \includegraphics[width=\linewidth]{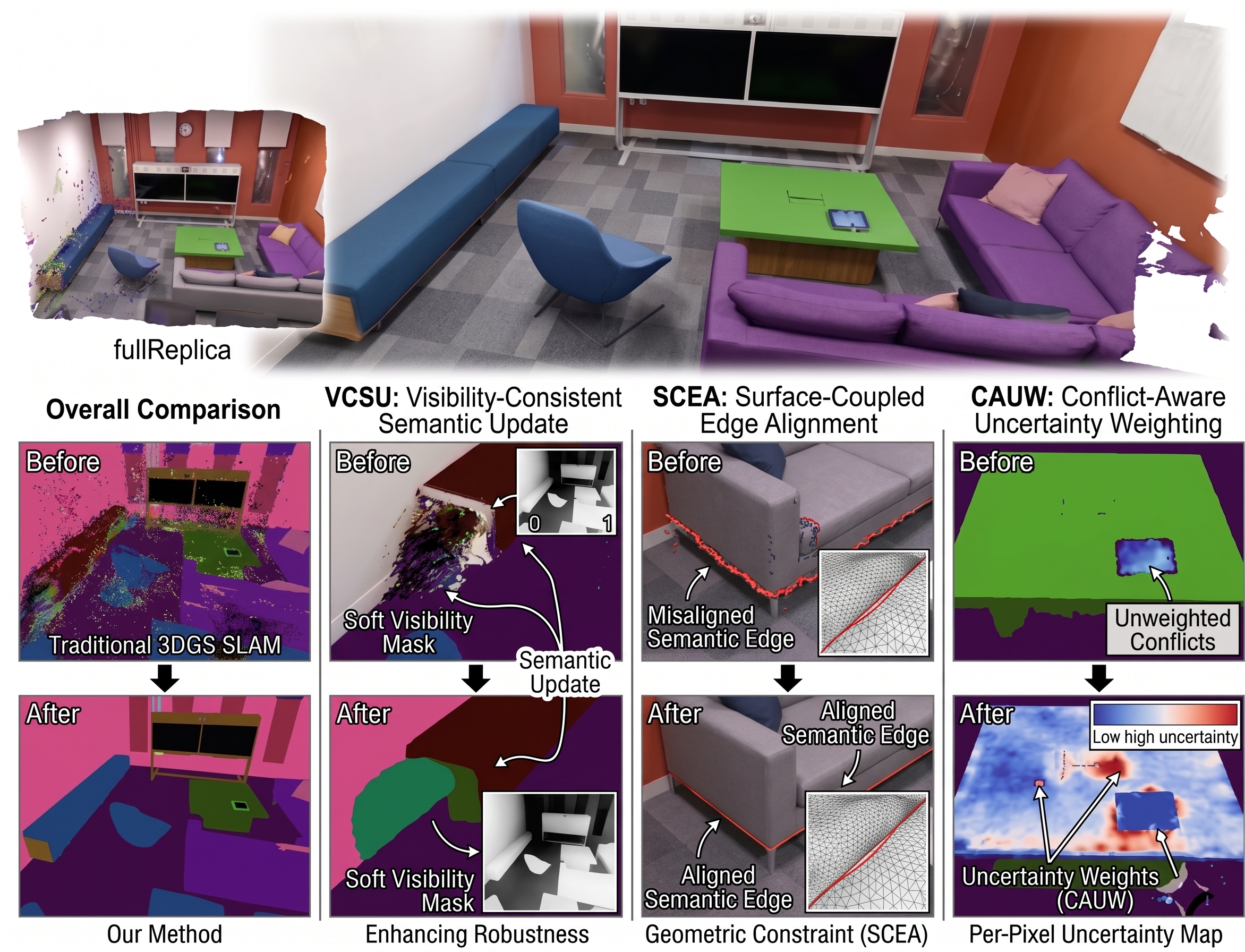}
    \vspace{-2mm}
    \caption{\textbf{Overview of VCS-SLAM.} Top: High-fidelity global semantic reconstruction. Bottom: Effectiveness of our modules against 3DGS baselines. (1) \textbf{Overall}: Enhanced global consistency. (2) \textbf{visibility-consistent semantic updates(VCSU)}: Mitigates occlusion artifacts via a depth-gated mask. (3) \textbf{Surface-coupled edge alignment(SCEA)}: Reduces unsupported semantic bleeding (4) \textbf{Conflict-aware uncertainty weighting (CAUW)}: Down-weights ambiguous observations via conflict-aware uncertainty mapping.}
    \vspace{-2mm}
    \label{fig:teaser}
    
\end{figure}

Recent breakthroughs in Neural Radiance Fields (NeRF) \cite{mildenhall2021nerf} and 3D Gaussian Splatting (3DGS) \cite{kerbl20233d} prompted their integration into visual SLAM, mitigating surface continuity limitations inherent in discrete mapping. Notably NeRF-based architectures (e.g. iMAP \cite{sucar2021imap} NICE-SLAM \cite{zhu2022nice}) and 3DGS frameworks (e.g. MonoGS \cite{matsuki2024gaussian} SplaTAM \cite{keetha2024splatam} SGS-SLAM \cite{li2024sgs}) achieve state-of-the-art performance producing consistent geometric and photorealistic maps.
Semantic mapping increasingly relies on 2D data-driven models. Lacking inherent spatial and temporal continuity, these models introduce severe semantic inconsistencies during online mapping. Accumulating spatial discrepancies significantly degrade semantic scene reconstruction accuracy. While recent frameworks \cite{zhu2025semgauss, yang2025opengs, li2024sgs} successfully integrate semantic features, they fail to adequately mitigate the propagation of these mapping errors. By overlooking visibility consistency, principled uncertainty weighting, and explicit coupling of semantic boundaries to geometric surfaces, current methods exhibit limited robustness in real-world applications.
Existing semantic 3D Gaussian SLAM methods treat per-frame labels as
uniformly reliable supervision, so geometrically inconsistent priors from occluded boundaries, smooth-surface bleeding, or ambiguous geometry are repeatedly fused into the map and become persistent artifacts. VCS-SLAM instead validates each observation's geometric reliability before fusion, preventing inconsistent updates from
accumulating in the Gaussian map.

To address these critical issues, this paper introduces a novel semantic SLAM framework featuring the following core contributions:

\begin{itemize}
    \item We formulate online semantic 3D Gaussian SLAM as a geometry-validated semantic evidence fusion problem, where 2D semantic priors are assimilated into the Gaussian map according to their geometric reliability rather than treated as uniformly valid supervision.

    \item We instantiate this reliability model using three complementary geometric cues: visibility-consistent semantic updates (VCSU), surface-coupled edge alignment (SCEA), and conflict-aware uncertainty weighting (CAUW). These cues jointly suppress occluded semantic updates, unsupported semantic bleeding, and premature label assignment in ambiguous regions.

    \item We integrate the proposed reliability model into a differentiable semantic Gaussian optimization objective and validate it through targeted analyses on depth-inconsistent pixels, semantic boundary regions, and uncertainty-error calibration, through a robustness study under structured corruption of the 2D priors, and against naive reweighting variants of each module.
\end{itemize}

\section{RELATED WORKS}

\subsection{Traditional Semantic SLAM-based methods}
Conventional semantic SLAM frameworks struggle to seamlessly integrate semantic information into mapping processes. Historically, these methods relied on discrete or implicit geometries like point clouds, voxels, and signed distance fields (SDFs) \cite{narita2019panopticfusion, 6907236, jha2025adaptive}. Point clouds and voxel grids are highly vulnerable to sparsity and degradation during rapid camera motion. Conversely, SDFs offer continuous surfaces but incur severe computational overhead, restricting real-time scalability. Consequently, a critical need remains for efficient continuous mapping methods providing robust semantic-geometric representations.

\subsection{SLAM with Neural Radiance Fields}
The remarkable performance of Neural Radiance Fields (NeRF) in high-fidelity novel view synthesis \cite{mildenhall2021nerf} motivated their integration into SLAM frameworks \cite{sucar2021imap, zhu2022nice}. For instance, SNI-SLAM \cite{zhu2024sni} employs hierarchical semantic encoding and cross-attention mechanisms fusing appearance, geometric, and semantic features. However, the computational overhead inherent in NeRF volume rendering makes it challenging for dense SLAM methods to simultaneously model and optimize environmental semantic maps in real time \cite{haghighi2023neural, li2024dns}. Additionally, the implicit nature of these representations fundamentally limits overall NeRF-based system performance \cite{tosi2026nerfs}.

\subsection{SLAM with Gaussian Splatting}
3D Gaussian Splatting (3DGS) has emerged as a promising representation utilizing 3D Gaussians parameterized by position, anisotropic covariance, opacity, and color \cite{kerbl20233d}. Existing 3DGS methods predominantly focus on RGB reconstruction, refining camera pose estimation via differentiable rendering with less attention devoted to semantic reconstruction \cite{matsuki2024gaussian, keetha2024splatam, yan2024gs, yugay2023gaussian}. Recent approaches integrate semantics by encoding per-frame labels as an auxiliary channel within Gaussian parameters \cite{li2024sgs, zhu2025semgauss, li2025hier}. Observation reliability, however, is addressed only indirectly: NEDS-SLAM \cite{ji2024neds} enforces consistency at the representation level, and OpenGS-SLAM \cite{yang2025opengs} handles open-set labels obtained from 2D predictors, but neither weights individual observations by whether the reconstructed geometry supports them. Where uncertainty weighting is used in semantic prediction, per-pixel confidence is typically derived from the predictor's own output distribution \cite{kendall2017uncertainties}. Such semantic-derived confidence is blind to the geometric state of the map: a confidently predicted label on an occluded or unconverged surface remains confidently wrong. VCS-SLAM instead derives reliability from rendered geometry, through depth residual, surface support, and ray-level depth variance, gating supervision by whether the map can currently support an observation rather than by how certain the 2D predictor is.

\section{METHODOLOGY}

\begin{figure*}[htbp]
    \centering
    \includegraphics[width=0.9\linewidth]{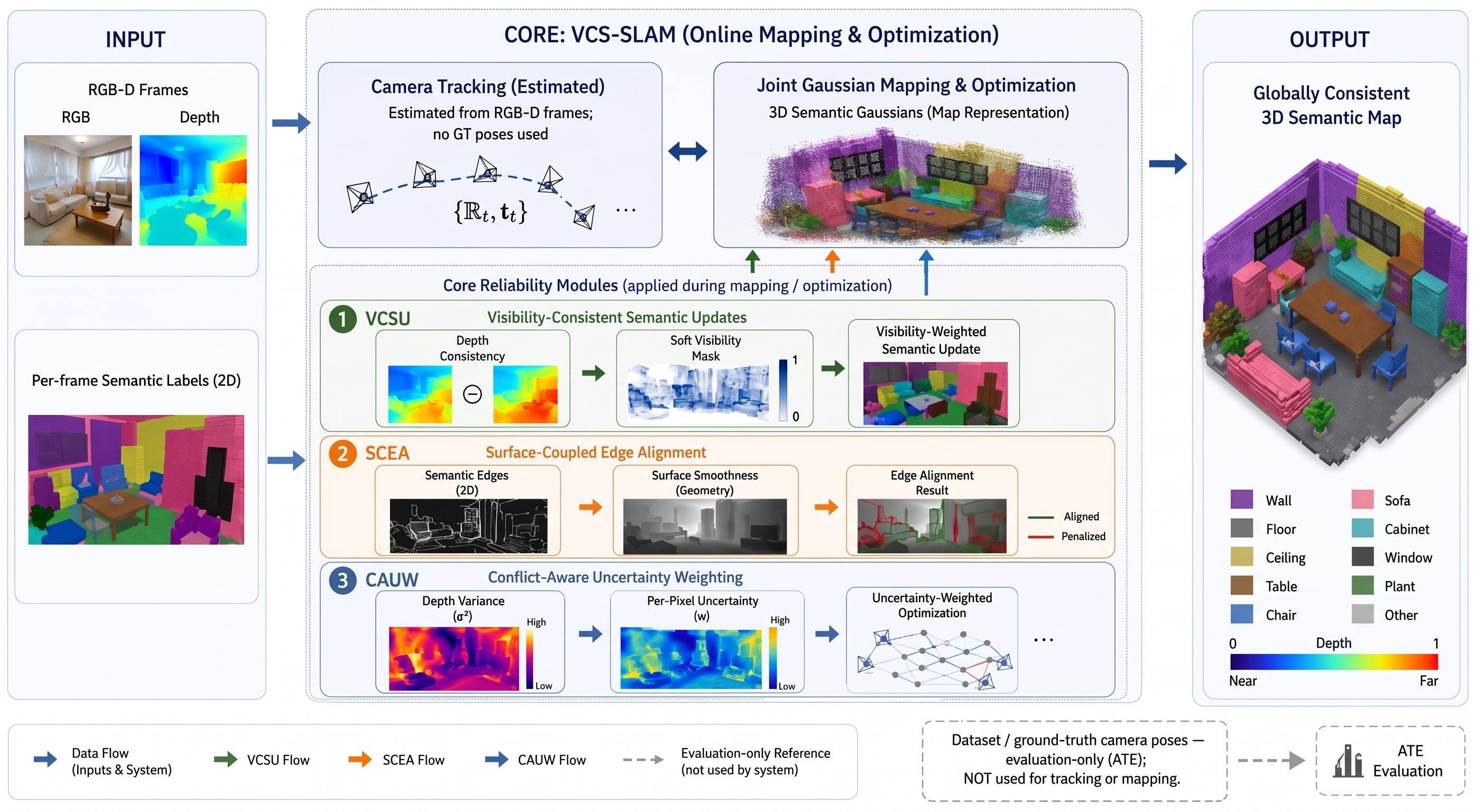}
    \vspace{-4mm}
    \caption{\textbf{System architecture of VCS-SLAM.} \textbf{Input:} RGB-D frames and per-frame 2D semantic labels. Camera poses are \emph{not} an input but are estimated online by the tracker from RGB-D alone; dataset ground-truth poses (dashed path) serve only as an ATE evaluation reference, not for tracking or mapping. \textbf{Core:} camera tracking and joint Gaussian mapping are optimized alternately over a shared set of 3D semantic Gaussians. Three reliability modules gate the semantic supervision during mapping: visibility-consistent semantic updates (VCSU) attenuate observations where rendered and sensor depth disagree; surface-coupled edge alignment (SCEA) penalizes semantic boundaries lacking a geometric edge; and conflict-aware uncertainty weighting (CAUW) down-weights pixels with high ray-level depth variance. \textbf{Output:} dense 3D semantic map and novel-view rendering.}
    \vspace{-2mm}
\label{fig:architecture}
\end{figure*}

\subsection{Semantic Scene Representation using Gaussian Splatting}
Our framework extends 3D Gaussian Splatting \cite{kerbl20233d} with semantic information. The environment is modeled as 3D Gaussians with position $\mu$, opacity $\sigma$, and geometric scale, each augmented with an appearance channel $c_i$ and a $C$-dimensional semantic logit vector $s_i$, where $C$ is the number of classes. Per-frame labels are converted to one-hot targets for pixel-wise cross-entropy supervision.

Gaussians are differentiably projected onto the image plane and aggregated by depth-sorted front-to-back volume rendering. Let $\alpha_i$ denote the evaluated 2D influence of a Gaussian including its occlusion term. The semantic map $S_{p}$ is rendered by blending the semantic vectors of contributing Gaussians:
\vspace{-3mm}
\begin{equation}
S_{p} = \sum_{i=1}^{n} s_i \alpha_i \prod_{j=1}^{i-1} (1 - \alpha_j)
\end{equation}
Similarly expected depth $D_{p}$ is synthesized by accumulating depth $d_i$ of each Gaussian center in camera coordinates:
\vspace{-3mm}
\begin{equation}
D_{p} = \sum_{i=1}^{n} d_i \alpha_i \prod_{j=1}^{i-1} (1 - \alpha_j).
\end{equation}

This baseline formulation applies gradients uniformly across the view, assuming all projected semantic features are equally reliable, and therefore cannot account for visibility inconsistencies, edge misalignment, or depth uncertainty. We address this with a visibility-consistent and surface-coupled mapping paradigm.

\subsection{Geometry Validated Semantic Evidence Fusion}

For each pixel $p$, the rendered semantic prediction $S_p$ is supervised by a 2D semantic prior $y_p$. Instead of applying this supervision uniformly, we define a geometric reliability score $\rho_p$ that measures whether the semantic observation is valid for updating the Gaussian map:
\vspace{-2mm}
\begin{equation}
\rho_p = M_{vis}(p) \cdot C_{p}
\end{equation}

where $M_{vis}(p)$ evaluates visibility consistency and $C_{p}$ evaluates ray level conflict confidence. Surface-supported boundary consistency is further imposed through a structural regularizer $\mathcal{L}_{edge}$. The semantic objective is written as:
\vspace{-2mm}
\begin{equation}
\mathcal{L}_{sem}
= \frac{1}{N}\sum_p \rho_p \cdot CE(S_p,y_p)
+ \lambda_{edge}\mathcal{L}_{edge}
\end{equation}

Under this unified formulation, VCSU provides $M_{vis}$, CAUW provides $C_{p}$, and SCEA provides $\mathcal{L}_{edge}$. This ensures 2D semantic priors are assimilated according to their geometric reliability rather than treated as uniformly valid supervision. Eq. 4 is a simplified form detailed in III-D.   

\subsection{Visibility-Consistent and Surface-Coupled Semantic Mapping}

\subsubsection{\textbf{Visibility-Consistent Semantic Updates}}

Conventional semantic mapping commonly relies on per-frame
semantic observations, which may come from dataset annotations
or 2D segmentation models. Although such observations provide
useful category supervision, they are fused frame by frame and
can become geometrically inconsistent under occlusion, boundary
misalignment, or incomplete reconstruction. Direct integration of these fluctuating observations corrupts the global semantic map. To mitigate this, frameworks like SNI-SLAM \cite{zhu2024sni} utilize cross-attention mechanisms fusing features. Similarly CoSegGaussians \cite{dou2024learning} embeds DINO \cite{caron2021emerging} features into Gaussian parameters, leveraging multi-view scale consistency.

While improving consistency these approaches treat all semantic observations equally during optimization, failing to reject erroneous updates from occluded boundaries. To address this we propose a depth-gated Visibility-Consistent Semantic Update (VCSU) module. Pixels exhibiting significant discrepancies between rendered expected depth and sensor depth correspond to occluded dynamic or poorly reconstructed regions, making their semantic observations unreliable.

To attenuate these unreliable updates we formulate a soft visibility mask $M_{vis}$ applying a Gaussian weighting function to the depth residual. Given the differentiably rendered depth $D_p$ and the observed input depth $D_p^{obs}$ visibility weight is:

\begin{equation}
M_{vis}(p) = \exp\left(-\frac{|D_p - D_p^{obs}|^2}{2\sigma_{vis}^2}\right)
\end{equation}

where $D_p^{obs}$ denotes the RGB-D input depth at pixel $p$ and $\sigma_{vis} = 0.05$ is a bandwidth hyperparameter dictating sensitivity to geometric inconsistencies. Acting as a dynamic depth gating mechanism during optimization $M_{vis}(p)$ is applied to the semantic loss calculation, down-weighting gradients at pixels suffering from depth misalignment.

\subsubsection{\textbf{Surface-Coupled Edge Alignment}}

A fundamental limitation of continuous volumetric representations is semantic feature bleeding across geometric boundaries. While VCSU mitigates visibility inconsistencies, it cannot prevent spatial oversmoothing at object edges. To address this, we introduce a Surface-Coupled Edge Alignment (SCEA) module. SCEA discourages unsupported semantic gradients in geometrically smooth and low-confidence regions, reducing semantic bleeding while limiting the penalty on transitions with partial geometric support.

We extract spatial gradients of rendered expected depth $D_{p}$ and semantic map $S_{p}$ using a Sobel operator. The edge magnitude $E$ for a 2D tensor $I$ is:
\vspace{-2mm}
\begin{equation}
E(I) = \sqrt{(\nabla_x * I)^2 + (\nabla_y * I)^2}
\end{equation}

where $\nabla_x$ and $\nabla_y$ are horizontal and vertical Sobel convolution kernels. Let $E_D = E(D_{p})$ and $E_S = E(S_{p})$ denote the depth and semantic edge maps. For our multi-channel representation, $E_S$ is the maximum edge magnitude across individual semantic channels capturing categorical transitions.

To formulate the boundary constraint, the depth edge map is normalized to $[0, 1]$ as a structural reference:
\vspace{-2mm}
\begin{equation}
\hat{E}_D = \frac{E_D}{\max(E_D) + \epsilon}
\end{equation}

where $\epsilon = 10^{-8}$ ensures numerical stability. 

Rather than enforcing a hard constraint that requires every semantic boundary to coincide with a depth discontinuity, we formulate a soft surface-support penalty. The surface-coupled edge loss $\mathcal{L}_{edge}$ evaluates semantic edge magnitudes weighted by the lack of geometric support:
\vspace{-2mm}
\begin{equation}
\mathcal{L}_{edge} = \frac{1}{N} \sum E_S \cdot \max(0, \tau_d - \hat{E}_D) 
\end{equation}

where $\tau_d \in [0,1]$ is a tolerance on the normalized depth-edge
response. A semantic edge is penalized in proportion to how far its
geometric support falls below $\tau_d$ and is left unpenalized once
$\hat{E}_D \geq \tau_d$; we set $\tau_d = 0.75 $. This coupling guides 3D Gaussian optimization to reduce isolated semantic bleeding at weakly supported boundaries.

\subsubsection{\textbf{Conflict-Aware Uncertainty Weighting}}

While previous modules enforce structural alignment and visibility, standard optimization suffers when observations originate from noisy or conflicting geometric regions. Treating all visible pixels equally degrades the semantic map, even when the underlying 3D geometry remains unconverged. We propose a Conflict-Aware Uncertainty Weighting (CAUW) mechanism to evaluate ray-level conflict confidence and modulate semantic residuals according to geometric reliability per pixel.

Within the volumetric rendering framework, we evaluate expected depth and its higher-order moments, extracting pixel-wise geometric uncertainty. Let expected depth be $E[D] = D_{p}$ and expected squared depth be $E[D^2]$. The depth variance $V_{pix}$ along the rendered ray is:
\vspace{-1mm}
\begin{equation}
V_{pix} = \max(E[D^2] - (E[D])^2, \epsilon)
\end{equation}

where $\epsilon = 10^{-8}$ ensures numerical stability. High depth variance indicates geometric conflicts (e.g., object boundaries or poorly optimized splats), making the corresponding semantic observation uncertain.

CAUW converts this ray-wise depth variance into a local semantic confidence score $C_{p}$:
\vspace{-1mm}
\begin{equation}
C_{p} = \frac{1}{1 + \gamma V_{pix}}
\end{equation}

where $\gamma = 100.0$ controls sensitivity to depth conflicts. Instead of applying a global image-level weight, we utilize $C_{p}$ to perform true per-pixel uncertainty weighting. Pixels with high depth variance, indicating competing or poorly optimized geometry along the ray, receive smaller semantic gradients. This prevents ambiguous regions from prematurely committing to incorrect semantic labels, improving consistency of the semantic reconstruction.

\subsection{Joint Optimization Objective}

We formulate a unified joint optimization objective to simultaneously optimize 3D Gaussian parameters across appearance, geometric, and semantic channels, explicitly accounting for our geometry-validated reliability model.

We modulate pixel-wise semantic residuals using the reliability score $\rho_p$ of Eq. 3, which combines the depth-gated mask $M_{vis}$ with the local confidence score $C_p$. To maintain stability, the semantic fidelity term interpolates the reliability-gated and standard cross-entropy (CE) losses per pixel:

\vspace{-1mm}
\begin{equation}
\mathcal{L}_{sem}^{fid} = \beta \frac{1}{N} \sum_p \rho_p \text{CE}(S_p, y_p) + (1 - \beta) \frac{1}{N} \sum_p \text{CE}(S_p, y_p)
\end{equation}

where $\beta = 0.7$ is a blending coefficient.

Incorporating SCEA, we formulate our final semantic loss $\mathcal{L}_{sem}^{ours}$. The structural boundary constraint $\mathcal{L}_{edge}$ acts as an additive penalty:

\vspace{-2mm}
\begin{equation}
\mathcal{L}_{sem}^{ours} = \mathcal{L}_{sem}^{fid} + \lambda_{edge} \mathcal{L}_{edge}
\end{equation}
\vspace{-2mm}

where $\lambda_{edge} = 0.02$ determines the geometric boundary coupling strength.

Finally, this loss is integrated into the overarching mapping objective. The complete joint mapping loss is:

\vspace{-2mm}
\begin{equation}
\mathcal{L}_{mapping} = \lambda_D \mathcal{L}_D + \lambda_C \mathcal{L}_C + \lambda_S \mathcal{L}_{sem}^{ours}
\end{equation}

where $\mathcal{L}_D$ and $\mathcal{L}_C$ represent standard depth and photometric rendering losses with their respective $\lambda$ balancing weights. Optimizing this objective ensures per-frame semantic priors are assimilated strictly according to their geometric reliability, yielding a more globally consistent semantic map. VCSU and CAUW target complementary failures: VCSU rejects occluded priors
through rendered-versus-sensor depth, while CAUW down-weights intra-ray
ambiguity flagged by high depth variance.

\begin{figure*}[htbp]
    \centering
    \includegraphics[width=0.95\linewidth]{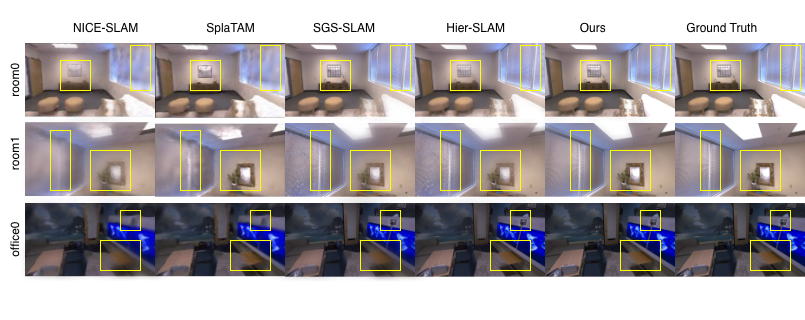}
    \vspace{-10mm}
    \caption{\textbf{Qualitative RGB reconstruction comparison on three Replica scenes \cite{straub2019replica} (Room-0, Room-1, Office-0)}. We compare renderings from NICE-SLAM \cite{zhu2022nice}, SplaTAM \cite{keetha2024splatam}, SGS-SLAM \cite{li2024sgs}, Hier-SLAM \cite{li2025hier}, and VCS-SLAM (Ours) against ground truth. Colored boxes mark thin structures and high-frequency regions where methods differ most. All panels, including baselines, are produced by our own runs under a common evaluation protocol. VCS-SLAM preserves fine geometric detail and sharper boundaries, whereas implicit reconstruction tends to oversmooth these regions.}
    \vspace{-2mm}
    \label{fig:comparison}
\end{figure*}

\section{EXPERIMENTS}

\subsection{Experimental Setup}

\subsubsection{\textbf{Dataset}}
Following standard dense SLAM protocols, we evaluate our framework
on synthetic and real RGB-D datasets. For synthetic evaluation, we
use eight scenes from Replica, which provide RGB-D frames,
per-pixel semantic labels rendered via Habitat-Sim, and ground-truth
camera poses for evaluation only. For real-world evaluation, we use
six scenes from the ScanNet, which provides RGB-D sequences, per-frame
2D semantic annotations and BundleFusion\cite{dai2017bundlefusion} camera poses were used as evaluation references. The SLAM system uses RGB-D frames and
per-frame semantic labels as input supervision; evaluation poses are
not used during tracking or mapping.

\subsubsection{\textbf{Metrics}}
We employ comprehensive metrics to quantitatively assess reconstruction fidelity and system robustness. Rendering quality is evaluated using PSNR, SSIM \cite{wang2004image}, and LPIPS \cite{zhang2018unreasonable}. Geometric accuracy is measured using the Depth L1 error on rendered 2D depth maps. We quantify camera tracking precision using absolute trajectory error \cite{sturm2012benchmark}. Finally 3D semantic mapping efficacy is evaluated using the mean intersection over Union (mIoU) metric.

\subsubsection{\textbf{Baselines}}
We benchmark our approach against state-of-the-art dense visual SLAM systems. For geometric reconstruction and camera tracking we compare against implicit neural methods \cite{zhu2022nice,wang2023co,johari2023eslam} and explicit 3DGS frameworks \cite{keetha2024splatam,zhu2025semgauss,li2025hier}. For dense semantic mapping evaluation, we compare against the NeRF-based semantic method SNI-SLAM \cite{zhu2024sni} alongside contemporary 3DGS semantic SLAM systems (SGS-SLAM \cite{li2024sgs}, SemGauss-SLAM \cite{zhu2025semgauss}, Hier-SLAM \cite{li2025hier}).

\subsubsection{\textbf{Implementation Details}}
VCS-SLAM is implemented in PyTorch and evaluated on an NVIDIA RTX 4090 GPU. The framework processes RGB-D frames online: camera tracking runs on every frame, while global mapping and Gaussian densification are triggered every  $N{=}8$ frames. 2D semantic priors are provided as per-frame label images, using
ground-truth semantic annotations on Replica and dataset-provided
semantic annotations on ScanNet. Each Gaussian stores a C-dimensional semantic logit vector, which is jointly optimized with appearance and geometry using pixel-wise cross-entropy supervision. Loss weighting coefficients balance multi-channel gradient contributions. For tracking, photometric and depth weights are $\lambda_{C}^{track}{=}0.5$ and $\lambda_{D}^{track}{=}1.0$. For mapping, these are maintained at $\lambda_{C}^{map}{=}0.5$ and $\lambda_{D}^{map}{=}1.0$ with semantic loss weight $\lambda_S{=}0.01$. Module-specific hyperparameters are: depth-gating bandwidth $\sigma_{vis}{=}0.05$ and blending coefficient $\beta{=}0.7$ (VCSU), depth-variance scaling $\gamma{=}100.0$ (CAUW), and boundary penalty $\lambda_{edge}{=}0.02$ with depth-edge support tolerance $\tau_d{=}0.75$ (SCEA). These values prevent gradient domination across feature channels during convergence.

\subsection{SLAM Performance}

\subsubsection{Tracking Performance}
We evaluate camera tracking accuracy on the synthetic Replica \cite{straub2019replica} and real ScanNet \cite{dai2017scannet} datasets, with results shown in Tab. \ref{tab:localization_rmse} and Tab. \ref{tab:scannet_rmse}, respectively. On Replica, our approach achieves the lowest average ATE RMSE among the compared methods. ScanNet poses greater challenges due to sensor noise and camera blur, leading to higher trajectory errors across all evaluated methods. Despite these dynamic conditions, our system maintains robust tracking capabilities alongside leading SOTA methods.

\begin{table}[htbp]
\centering
\caption{\textsc{Localization performance ATE RMSE (cm) on the Replica dataset. Best results are highlighted as \colorbox{green!25}{\textbf{FIRST}}, \colorbox{yellow!25}{SECOND}.}}
\label{tab:localization_rmse}
\resizebox{\columnwidth}{!}{%
\begin{tabular}{@{} l ccccccccc @{}}
\toprule
\textbf{Methods} & \textbf{Avg.$\downarrow$} & \textbf{R0$\downarrow$} & \textbf{R1$\downarrow$} & \textbf{R2$\downarrow$} & \textbf{Of0$\downarrow$} & \textbf{Of1$\downarrow$} & \textbf{Of2$\downarrow$} & \textbf{Of3$\downarrow$} & \textbf{Of4$\downarrow$} \\
\midrule
iMap \cite{sucar2021imap}            & 4.15 & 6.33 & 3.46 & 2.65 & 3.31 & 1.42 & 7.17 & 6.32 & 2.55 \\
NICE-SLAM \cite{zhu2022nice}       & 1.07 & 0.97 & 1.31 & 1.07 & 0.88 & 1.00 & 1.06 & 1.10 & 1.13 \\
Vox-Fusion \cite{yang2022vox}      & 3.09 & 1.37 & 4.70 & 1.47 & 8.48 & 2.04 & 2.58 & 1.11 & 2.94 \\
co-SLAM \cite{wang2023co}         & 1.06 & 0.72 & 0.85 & 1.02 & 0.69 & 0.56 & 2.12 & 1.62 & 0.87 \\
ESLAM \cite{johari2023eslam}            & 0.63 & 0.71 & 0.70 & 0.52 & 0.57 & 0.55 & 0.58 & 0.72 & 0.63 \\
Point-SLAM \cite{sandstrom2023point}      & 0.52 & 0.61 & 0.41 & 0.37 & 0.38 & 0.48 & 0.54 & 0.69 & 0.72 \\
MonoGS \cite{matsuki2024gaussian}          & 0.79 & 0.47 & 0.43 & 0.31 & 0.70 & 0.57 & 0.31 & \cellcolor{green!25}0.31 & 3.2 \\
SplaTAM \cite{keetha2024splatam}          & 0.36 & 0.31 & \cellcolor{yellow!25}0.40 & 0.29 & 0.47 & 0.27 & \cellcolor{yellow!25}0.29 & \cellcolor{yellow!25}0.32 & 0.55 \\
SNI-SLAM \cite{zhu2024sni}         & 0.46 & 0.50 & 0.55 & 0.45 & 0.35 & 0.41 & 0.33 & 0.62 & \cellcolor{yellow!25}0.50 \\
DNS SLAM \cite{li2024dns}         & 0.45 & 0.49 & 0.46 & 0.38 & 0.34 & 0.35 & 0.39 & 0.62 & 0.60 \\
SemGauss-SLAM \cite{zhu2025semgauss}   & \cellcolor{yellow!25}0.33 & 0.26 & 0.42 & \cellcolor{yellow!25}0.27 & 0.34 & \cellcolor{yellow!25}0.17 & 0.32 & 0.36 & \cellcolor{green!25}0.49 \\
Hier-SLAM \cite{li2025hier}  & \cellcolor{yellow!25}0.33 & \cellcolor{green!25}0.21 & 0.49 & \cellcolor{green!25}0.24 & \cellcolor{yellow!25}0.29 & \cellcolor{green!25}0.16 & 0.31 & 0.37 & 0.53 \\
\textbf{VCS-SLAM (Ours)}  & \cellcolor{green!25}\textbf{0.30} & \cellcolor{yellow!25}\textbf{0.23} & \cellcolor{green!25}\textbf{0.39} & \cellcolor{green!25}\textbf{0.24} & \cellcolor{green!25}\textbf{0.28} & \cellcolor{yellow!25}\textbf{0.17} & \cellcolor{green!25}\textbf{0.28} & \cellcolor{yellow!25}\textbf{0.32} & \textbf{0.51} \\
\bottomrule
\end{tabular}%
}
\end{table}

\begin{table}[htbp]
\centering
\caption{\textsc{Localization performance ATE RMSE (cm) on the Scannet dataset. Best results are highlighted as \colorbox{green!25}{FIRST}, \colorbox{yellow!25}{SECOND}, \colorbox{red!25}{THIRD}.}}
\label{tab:scannet_rmse}
\resizebox{\columnwidth}{!}{%
\begin{tabular}{@{} l ccccccc @{}}
\toprule
\textbf{Methods} & \textbf{Avg.$\downarrow$} & \textbf{0000$\downarrow$} & \textbf{0059$\downarrow$} & \textbf{0106$\downarrow$} & \textbf{0169$\downarrow$} & \textbf{0181$\downarrow$} & \textbf{0207$\downarrow$} \\
\midrule
NICE-SLAM \cite{zhu2022nice}       & \cellcolor{yellow!25}10.70 & 12.00 & 14.00 & \cellcolor{green!25}7.90 & \cellcolor{red!25}10.90 & 13.40 & \cellcolor{green!25}6.20 \\
Vox-Fusion \cite{yang2022vox}      & 26.90 & 68.84 & 24.18 & \cellcolor{yellow!25}8.41 & 27.28 & 23.30 & 9.41 \\
Point-SLAM \cite{sandstrom2023point}      & 12.19 & \cellcolor{green!25}10.24 & \cellcolor{green!25}7.81 & \cellcolor{red!25}8.65 & 22.16 & 14.77 & 9.54 \\
SplaTAM \cite{keetha2024splatam}          & 11.88 & 12.83 & 10.10 & 17.72 & 12.08 & 11.10 & \cellcolor{red!25}7.46 \\
SemGauss-SLAM \cite{zhu2025semgauss}   & -- & 11.87 & \cellcolor{yellow!25}7.97 & -- & \cellcolor{green!25}8.70 & \cellcolor{yellow!25}9.78 & 8.97 \\
Hier-SLAM \cite{li2025hier}    & \cellcolor{red!25}11.36 & \cellcolor{red!25}11.45 & \cellcolor{red!25}9.61 & 17.80 & 11.93 & \cellcolor{red!25}10.04 & \cellcolor{yellow!25}7.32 \\
\textbf{VCS-SLAM (Ours)}     & \cellcolor{green!25}\textbf{10.22} & \cellcolor{yellow!25}\textbf{10.85} & \textbf{12.36} & \textbf{9.23} & \cellcolor{yellow!25}\textbf{10.51} & \cellcolor{green!25}\textbf{9.75} & \textbf{8.62} \\
\bottomrule
\end{tabular}%
}
\end{table}

\subsubsection{Mapping and Runtime Performance}
We evaluate mapping performance and geometric fidelity using the Depth L1 error metric on the Replica dataset. Tab. \ref{tab:mapping_runtime} shows that our proposed methodology achieves the lowest depth estimation error of 0.321 cm. These results are consistent with the effect of our spatial consistency constraints. Furthermore, we quantify system runtime to validate operational efficiency. Despite integrating complex reliability modules, VCS-SLAM maintains a highly competitive overall SLAM speed of 7.27 FPS. Baseline results are taken from \cite{li2024sgs}, VCS-SLAM results are from our implementation.

\begin{table}[htbp]
\centering
\caption{\textsc{Mapping and Runtime Performance Comparison on the Replica Dataset} Best results are highlighted as \colorbox{green!25}{\textbf{FIRST}}, \colorbox{yellow!25}{SECOND}}
\label{tab:mapping_runtime}
\resizebox{\columnwidth}{!}{%
\begin{tabular}{lcccc}
\toprule
\textbf{Methods} &
  \begin{tabular}[c]{@{}c@{}}Depth L1\\ {[}cm{]}$\downarrow$\end{tabular} &
  \begin{tabular}[c]{@{}c@{}}Track. FPS\\ {[}f/s{]}$\uparrow$\end{tabular} &
  \begin{tabular}[c]{@{}c@{}}Map. FPS\\ {[}f/s{]}$\uparrow$\end{tabular} &
  \begin{tabular}[c]{@{}c@{}}SLAM FPS\\ {[}f/s{]}$\uparrow$\end{tabular} \\
\midrule
NICE-SLAM \cite{zhu2022nice}     & 1.903 & 13.70 & 0.20 & 0.20 \\
Co-SLAM \cite{wang2023co}      & 1.513 & 17.24 & \cellcolor{green!25}{10.20} & \cellcolor{yellow!25}{6.41}  \\
ESLAM \cite{johari2023eslam}        & 1.180 & \cellcolor{green!25}{18.11}  & 3.62 & 3.02 \\
SplaTAM \cite{keetha2024splatam}      & 0.525 & 5.53 & 3.84 & 2.26  \\
SGS-SLAM \cite{li2024sgs} & \cellcolor{yellow!25}{0.356} & 5.27 & 3.52 & 2.11  \\
\textbf{VCS-SLAM(Ours)} & \cellcolor{green!25}\textbf{0.321} & \cellcolor{yellow!25}\textbf{17.71} & \cellcolor{yellow!25}\textbf{9.56} & \cellcolor{green!25}\textbf{7.27}  \\
\bottomrule
\end{tabular}%
}
\end{table}

\subsubsection{Rendering Performance}
Following established protocols \cite{zhu2022nice, sandstrom2023point, keetha2024splatam, li2024sgs, li2025hier}, we evaluate rendering quality across eight Replica \cite{straub2019replica} scenes. As shown in Tab. \ref{tab:rendering_replica}, VCS-SLAM attains the best average PSNR, SSIM, and LPIPS among the compared methods. The results are reported from \cite{li2025hier}, VCS-SLAM results are ours.

\begin{table}[htbp]
\centering
\caption{Rendering performance on Replica, averaged over eight scenes.
Best Results are highlighted as
\colorbox{green!25}{FIRST}, \colorbox{yellow!25}{SECOND}.}
\label{tab:rendering_replica}
\resizebox{\columnwidth}{!}{%
\begin{tabular}{@{} l ccc @{}}
\toprule
\textbf{Methods} & \textbf{PSNR} $\uparrow$ & \textbf{SSIM} $\uparrow$ & \textbf{LPIPS} $\downarrow$ \\
\midrule
\multicolumn{4}{c}{\textbf{Visual SLAM}} \\
NICE-SLAM \cite{zhu2022nice}         & 24.42 & 0.809 & 0.233 \\
Co-SLAM \cite{wang2023co}            & 30.24 & 0.939 & 0.252 \\
ESLAM \cite{johari2023eslam}         & 29.08 & 0.929 & 0.239 \\
SplaTAM \cite{keetha2024splatam}     & 34.11 & 0.968 & 0.102 \\
\midrule
\multicolumn{4}{c}{\textbf{Semantic SLAM}} \\
SNI-SLAM \cite{zhu2024sni}           & 29.43 & 0.921 & 0.237 \\
SGS-SLAM \cite{li2024sgs}            & 34.66 & 0.973 & 0.096 \\
SemGauss-SLAM \cite{zhu2025semgauss} & 35.03 & \cellcolor{yellow!25}0.982 & \cellcolor{yellow!25}0.062 \\
Hier-SLAM \cite{li2025hier}          & \cellcolor{yellow!25}35.70 & 0.980 & 0.067 \\
\textbf{VCS-SLAM (Ours)}             & \cellcolor{green!25}\textbf{36.03} & \cellcolor{green!25}\textbf{0.984} & \cellcolor{green!25}\textbf{0.059} \\
\bottomrule
\end{tabular}%
}
\end{table}

\subsubsection{Semantic Performance}
Following the evaluation protocol of recent Gaussian-based semantic SLAM methods \cite{li2024sgs, zhu2025semgauss, li2025hier}, we benchmark semantic segmentation accuracy under the per-frame visible-class (subset) protocol on four Replica \cite{straub2019replica} scenes. We report under this single protocol because the directly comparable methods, SGS-SLAM, SemGauss-SLAM, and Hier-SLAM are evaluated under it, yielding a consistent and interpretable comparison. As shown in Tab.~\ref{tab:semantic_miou}, VCS-SLAM attains 95.87\% mean mIoU, second only to SemGauss-SLAM \cite{zhu2025semgauss} which leverages pre-trained foundation-model semantic supervision and ahead of Hier-SLAM \cite{li2025hier}, while achieving the highest accuracy on Room-0. These results show that the fusion of geometry-validated evidence delivers competitive global semantic accuracy, with the targeted contributions per-module isolated in Tab.~\ref{tab:mechanism_analysis}.

\begin{table}[htbp]
\centering
\caption{Semantic performance mIoU (\%) on Replica \cite{straub2019replica}. Best and second-best per column are highlighted as \colorbox{green!25}{\textbf{FIRST}} and \colorbox{yellow!25}{SECOND}.}
\label{tab:semantic_miou}
\resizebox{\columnwidth}{!}{%
\begin{tabular}{l|ccccc}
\toprule
\textbf{Methods} & \textbf{Avg.$\uparrow$} & \textbf{R0$\uparrow$} & \textbf{R1$\uparrow$} & \textbf{R2$\uparrow$} & \textbf{Of0$\uparrow$} \\
\midrule
SNI-SLAM \cite{zhu2024sni}      & 87.41 & 88.42 & 87.43 & 86.16 & 87.63 \\
SGS-SLAM \cite{li2024sgs}       & 92.72 & 92.95 & 92.91 & 92.10 & 92.90 \\
SemGauss-SLAM \cite{zhu2025semgauss} & \cellcolor{green!25}96.34 & \cellcolor{yellow!25}96.30 & \cellcolor{green!25}95.82 & \cellcolor{green!25}96.51 & \cellcolor{green!25}96.72 \\
Hier-SLAM \cite{li2025hier}     & 95.58 & 95.25 & \cellcolor{yellow!25}95.81 & \cellcolor{yellow!25}95.73 & 95.52 \\
\textbf{VCS-SLAM (Ours)}        & \cellcolor{yellow!25}\textbf{95.87} & \cellcolor{green!25}\textbf{96.44} & \textbf{95.78} & \textbf{95.63} & \cellcolor{yellow!25}\textbf{95.63} \\
\bottomrule
\end{tabular}%
}
\end{table}

\subsection{Qualitative Analysis}
Fig. \ref{fig:comparison} compares RGB reconstruction quality on Room-0, Room-1, and Office-0, with colored boxes marking thin structures and high-frequency texture where reconstruction methods diverge most. The implicit NICE-SLAM \cite{zhu2022nice} oversmooths these regions, rounding off depth transitions so that small or thin objects blur into the surrounding surface. Explicit Gaussian methods (SplaTAM \cite{keetha2024splatam}, SGS-SLAM \cite{li2024sgs}, Hier-SLAM \cite{li2025hier}) recover sharper local geometry, but residual blur and floaters persist on weakly observed or low-texture surfaces. VCS-SLAM yields the most consistent reconstructions in these regions: although its reliability modules act on semantic supervision, suppressing unreliable and conflicting gradients stabilizes densification and local geometry refinement, which in turn sharpens rendered structure and object boundaries. The semantic-specific effects of VCSU, SCEA, and CAUW, boundary preservation and reduced label bleeding are isolated separately in Fig. \ref{fig:ablation_visual} and Tab. \ref{tab:mechanism_analysis}.

\subsection{Ablation Study}

\begin{table}[htbp]
\centering
\caption{Ablation study on the Replica dataset}
\label{tab:ablation}
\resizebox{\ifdim\width>\columnwidth\columnwidth\else\width\fi}{!}{
\begin{tabular}{lcccc}
\toprule
\textbf{Settings} &
  \begin{tabular}[c]{@{}c@{}}Depth L1\\ {[}cm{]}$\downarrow$\end{tabular} &
  \begin{tabular}[c]{@{}c@{}}ATE RMSE\\ {[}cm{]}$\downarrow$\end{tabular} &
  \begin{tabular}[c]{@{}c@{}}PSNR\\ {[}dB{]}$\uparrow$\end{tabular} &
  \begin{tabular}[c]{@{}c@{}}mIoU\\ {[}\%{]}$\uparrow$\end{tabular} \\ 
\midrule
without VCSU module         & 0.60          & 0.65           & 32.51     & 91.23          \\
without SCEA module            & 0.53         & 0.79           & 33.47         & 93.62          \\
without CAUW module         & 0.54          & 0.70           & 34.21         & 91.24      \\
\midrule
With All modules                 & \textbf{0.31} & \textbf{0.29}  & \textbf{35.95}& \textbf{95.87} \\ 
\bottomrule
\end{tabular}}
\end{table}

\begin{figure}[htbp]
    \centering
    \includegraphics[width=\linewidth]{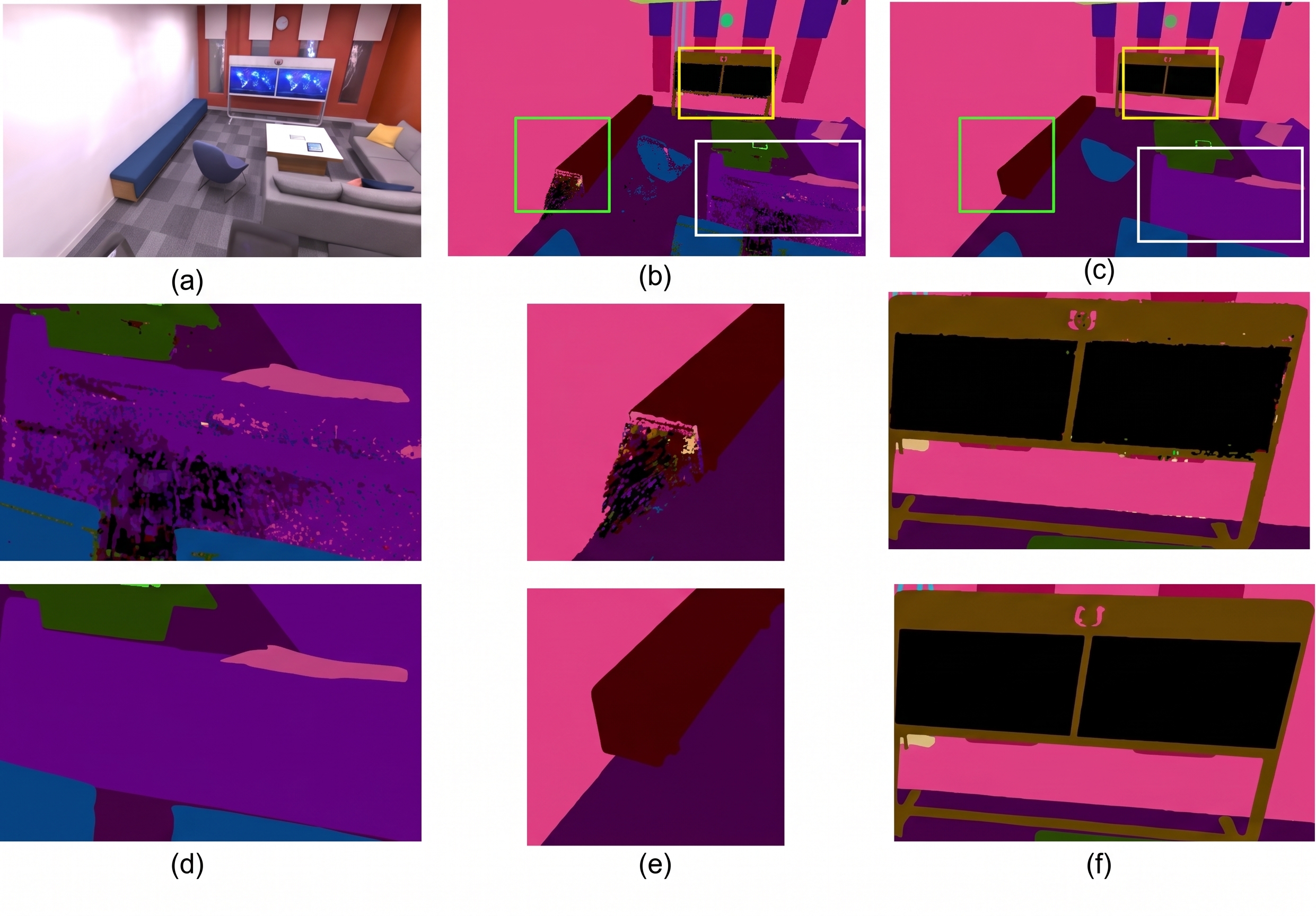}
    \vspace{-7mm}
    \caption{\textbf{Qualitative ablation study.} Image (a) shows the input RGB view; (b) our base 3DGS-SLAM model without VCSU, SCEA, and CAUW; (c) the full VCS-SLAM.}
    \vspace{-2mm}
    \label{fig:ablation_visual}
\end{figure}

Tab.~\ref{tab:ablation} reports each module's contribution on Replica, and Fig.~\ref{fig:ablation_visual} shows the corresponding qualitative effect. Removing any module degrades all four metrics: although the modules act only on semantic supervision, incorrect semantic gradients alter densification and local geometry refinement, so suppressing them also improves rendered depth and pose accuracy. Varying $\sigma_{vis}\!\in\!\{0.025, 0.05, 0.1\}$, $\gamma\!\in\!\{50, 100, 200\}$, and $\tau_d\!\in\!\{0.5, 0.75, 0.9\}$ one at a time on Room-0 changes mIoU by at most 1.23 points.

\vspace{-6pt}

\subsection{Mechanism Analysis}
Beyond global metrics, we evaluate targeted claims for each module in Tab.~\ref{tab:mechanism_analysis}. SGS-SLAM \cite{li2024sgs} is re-evaluated under our implementation using the same targeted-metric protocol (depth-inconsistent mIoU, boundary F1, error-detection AUROC), so all values in this table are our own.

\begin{table}[htbp]
\centering
\caption{Mechanism analysis of the core modules on Replica.}
\label{tab:mechanism_analysis}
\resizebox{\ifdim\width>\columnwidth\columnwidth\else\width\fi}{!}{%
\resizebox{\columnwidth}{!}{%
\begin{tabular}{@{}llccc@{}}
\toprule
\textbf{Module} & \textbf{Targeted Metric} & \textbf{SGS-SLAM} & \textbf{Ours} & \textbf{Improvement} \\
\midrule
\textbf{VCSU} & Depth-Inconsistent mIoU (\%) $\uparrow$ & 86.43 & \textbf{91.05} & \textbf{+4.62} \\
\textbf{SCEA} & Boundary F1 Score (BF) $\uparrow$ & 0.82 & \textbf{0.98} & \textbf{+0.16} \\
\textbf{CAUW} & Error Detection AUROC $\uparrow$ & 0.73 & \textbf{0.85} & \textbf{+0.12} \\
\bottomrule
\end{tabular}
}}
\end{table}

\textbf{Occlusion Robustness (VCSU):} We evaluate mIoU strictly on depth-inconsistent pixels, defined as those where rendered and sensor depth differ by more than 0.05 m. The 4.62\% absolute gain is consistent with the intended effect of the depth-gated visibility mask.

\textbf{Boundary Preservation (SCEA):} We compute Boundary F1 within a 3-pixel tolerance band around ground-truth morphological edges. The 0.16 increase indicates that surface-coupled constraints preserve crisper label transitions.

\textbf{Uncertainty Calibration (CAUW):} We correlate per-pixel uncertainty with semantic error, classifying a pixel as erroneous when its predicted class differs from ground truth. The 0.12 AUROC gain indicates that the proposed weighting better identifies ambiguous geometric regions.

\subsection{Robustness to Unreliable Semantic Priors}
\label{sec:robustness}

\begin{table}[htbp]
\centering
\caption{Noise Robustness}
\label{tab:noise_robustness}
\resizebox{\columnwidth}{!}{%
\begin{tabular}{@{} l l cccc c @{}}
\toprule
 & & \multicolumn{4}{c}{\textbf{mIoU (\%)} $\uparrow$} & \\
\cmidrule(lr){3-6}
\textbf{Corruption} & \textbf{Method} & \textbf{Clean} & \textbf{Mild} & \textbf{Mod.} & \textbf{Severe} & \textbf{$\Delta$} $\downarrow$ \\
\midrule
\multirow{2}{*}{Boundary ($k{=}1/3/5$ px)}
 & SGS-SLAM \cite{li2024sgs} & 92.72 & 90.17 & 86.49 & 82.37 & 10.35 \\
 & \textbf{VCS-SLAM (Ours)}  & \textbf{95.87} & \textbf{93.65} & \textbf{91.33} & \textbf{89.92} & \textbf{5.95} \\
\midrule
\multirow{2}{*}{Class confusion ($p{=}5/10/20\%$)}
 & SGS-SLAM \cite{li2024sgs} & 92.72 & 89.29 & 85.67 & 80.21 & 12.51 \\
 & \textbf{VCS-SLAM (Ours)}  & \textbf{95.87} & \textbf{93.07} & \textbf{90.77} & \textbf{88.91} & \textbf{6.96} \\
\midrule
\multirow{2}{*}{Missed detections ($q{=}10/25/50\%$)}
 & SGS-SLAM \cite{li2024sgs} & 92.72 & 91.03 & 87.52 & 83.61 & 9.11 \\
 & \textbf{VCS-SLAM (Ours)}  & \textbf{95.87} & \textbf{94.32} & \textbf{92.87} & \textbf{90.43} & \textbf{5.44} \\
\midrule
\multirow{2}{*}{Uniform, control ($r{=}5/10/20\%$)}
 & SGS-SLAM \cite{li2024sgs} & 92.72 & 90.37 & 87.95 & 85.38 & \textbf{7.34} \\
 & \textbf{VCS-SLAM (Ours)}  & \textbf{95.87} & \textbf{92.36} & \textbf{88.87} & \textbf{87.62} & 8.25 \\
\bottomrule
\end{tabular}%
}
\end{table}

The evaluations above draw labels from dataset annotations, under which every observation is already correct, and a reliability model has little to arbitrate. We therefore corrupt the 2D priors before fusion, reproducing the dominant failure modes of 2D segmentation: boundary misalignment (morphological dilation/erosion, $k{=}1,3,5$ px), class confusion between visually similar categories ($p{=}5,10,20\%$), and missed detections of small objects ($q{=}10,25,50\%$), with uniform random reassignment ($r{=}5,10,20\%$) as a control. As shown in Tab.~\ref{tab:noise_robustness}, VCS-SLAM degrades at little more than half the rate of SGS-SLAM \cite{li2024sgs} under all three structured corruptions, widening the margin from 3.15 mIoU on clean labels to between 6.82 and 8.70 at the most severe level. Under uniform reassignment, the trend reverses, which delimits the mechanism: the gates measure geometric support rather than semantic correctness, so an incorrect label at a geometrically consistent pixel is fused as readily by our method as by any other. 

\begin{table}[htbp]
\centering
\caption{Naive reweighting baselines on Replica. 
Averaged over the four scenes of Tab.~\ref{tab:semantic_miou}.
\emph{Struct.} is boundary misalignment at $k{=}3$ px. }
\label{tab:reweighting_baselines}
\resizebox{\columnwidth}{!}{%
\begin{tabular}{@{} l l cc c @{}}
\toprule
 & & \multicolumn{2}{c}{\textbf{mIoU (\%)} $\uparrow$} & \\
\cmidrule(lr){3-4}
\textbf{Variant} & \textbf{Substitution} & \textbf{Clean} & \textbf{Struct.} & \textbf{$\Delta$} $\downarrow$ \\
\midrule
Hard gate         & $M_{vis} \rightarrow \mathbb{1}[|D_p - D_p^{obs}| < \sigma_{vis}]$ & 92.23 & 88.12 & 4.11 \\
TV smoothness     & $\mathcal{L}_{edge} \rightarrow$ isotropic TV on $S_p$             & 92.19 & 87.68 & 4.51 \\
Entropy weighting & $C_p \rightarrow 1/(1 + H(\mathrm{softmax}(S_p)))$                 & 91.41 & 86.74 & 4.67 \\
\midrule
\textbf{VCS-SLAM (Ours)} & --- & \textbf{95.87} & \textbf{91.33} & \textbf{4.54} \\
\bottomrule
\end{tabular}%
}
\end{table}

To test whether these gains follow from the proposed formulation rather than from reweighting in general, we replace each module with a simpler counterpart of the same functional role and retrain, evaluating under clean labels and boundary corruption at $k{=}3$ px (Tab.~\ref{tab:reweighting_baselines}). All three variants degrade at a comparable rate to the full model, so the separation lies in accuracy level rather than robustness slope: each substitution costs 3.6--4.5 mIoU under clean labels and 3.2--4.6 mIoU under corruption. The entropy variant is the most direct comparison, deriving per-pixel confidence from the rendered semantic distribution instead of from geometry, and is the weakest of the three; a semantic distribution can be confidently peaked on an occluded or unconverged surface, whereas ray-level depth variance flags precisely that condition.

\section{CONCLUSION AND LIMITATION}

We propose a visibility-consistent, surface-coupled semantic mapping framework for RGB-D 3D Gaussian SLAM. Depth-gated visibility masking, surface-supported edge alignment, and conflict-aware uncertainty weighting validate per-frame semantic supervision before fusion into the persistent Gaussian map, reducing occlusion artifacts, unsupported semantic bleeding, and premature labeling in ambiguous regions. Experiments on Replica show improved semantic consistency, boundary preservation, rendering quality, and geometric reconstruction, with competitive tracking on ScanNet. Under structured corruption of the 2D priors, VCS-SLAM degrades at roughly half the rate of SGS-SLAM and outperforms simpler reweighting variants. Because reliability is derived from geometry, a depth residual can arise from pose error, incomplete geometry, or sensor noise as well as from occlusion, and semantic boundaries between coplanar regions receive no surface support. Validation against a live open-vocabulary segmenter and extension to open-set and dynamic scenes remain open.

\section*{ACKNOWLEDGMENT}
Authors appreciate the support provided by the NYUAD Center for Artificial Intelligence and Robotics (CAIR), funded by Tamkeen under the NYUAD Research Institute Award CG010.

\bibliographystyle{IEEEtran}
\bibliography{root}

\end{document}